\documentclass[10pt, a4paper]{article}

\usepackage{multirow}
\usepackage[table]{xcolor}
\usepackage{booktabs}
\usepackage{amssymb}
\usepackage{fancyvrb}

\usepackage[final]{lrec2026} 

\title{Toward Conversational Hungarian Speech Recognition: Introducing the BEA-Large and BEA-Dialogue Datasets}

\name{%
  \begin{minipage}{0.95\textwidth}
    \centering\bfseries
    Máté Gedeon$^{*,\dagger}$, Piroska Zsófia Barta$^{*,\dagger}$,
    Péter Mihajlik$^{*,\ddagger}$, Tekla Etelka Gráczi$^{\ddagger}$,
    Anna Kohári$^{\ddagger}$, Katalin Mády$^{\ddagger}$
    \vspace{8pt}
  \end{minipage}%
}
\vspace{15pt}
\address{%
  \begin{minipage}{0.9\textwidth}
    \centering
    $^{*}$Dept. of Telecommunications and Artificial Intelligence,\\
    Budapest University of Technology and Economics, Hungary\\
    $^{\ddagger}$ELTE Research Centre for Linguistics, Hungary\\
    $^{\dagger}$Speechtex Ltd.\\[4pt]
    \texttt{\{gedeonm, piroskazsofia.barta\}@edu.bme.hu}\\
    \texttt{\{mihajlik.peter, graczi.tekla.etelka, kohari.anna, mady\}@nytud.hu}
    \vspace{5pt}
  \end{minipage}
}

\abstract{
The advancement of automatic speech recognition (ASR) has been largely enhanced by extensive datasets in high-resource languages, while languages such as Hungarian remain underrepresented due to limited spontaneous and conversational corpora. To address this gap, we introduce two new datasets -- BEA-Large and BEA-Dialogue -- constructed from the previously unprocessed portions of the Hungarian speech corpus named BEA. BEA-Large extends BEA-Base with 255 hours of spontaneous speech from 433 speakers, enriched with detailed segment-level metadata. BEA-Dialogue, comprising 85 hours of spontaneous conversations, is a Hungarian speech corpus featuring natural dialogues partitioned into speaker-independent subsets, supporting research in conversational ASR and speaker diarization. We establish reproducible baselines on these datasets using publicly available ASR models, with the fine-tuned Fast Conformer model achieving word error rates as low as 14.18\% on spontaneous and 4.8\% on repeated speech. Diarization experiments yield diarization error rates between 12.46\% and 17.40\%, providing reference points for future improvements. The results highlight the persistent difficulty of conversational ASR, particularly due to disfluencies, overlaps, and informal speech patterns. By releasing these datasets and baselines, we aim to advance Hungarian speech technology and offer a methodological framework for developing spontaneous and conversational benchmarks in other languages.
 \\ \newline  }

\begin{document}

\maketitleabstract

\section{Introduction}
The field of automatic speech recognition (ASR) has been fundamentally transformed by the availability of large-scale training datasets \citelanguageresource{Librispeech, MLS_dataset}. However, this revolution has predominantly benefited well-resourced languages, leaving low-resource languages like Hungarian significantly underrepresented in the modern ASR ecosystem \cite{Low_resource_survey}. While recent advances in self-supervised learning and multilingual models have shown promise for low-resource scenarios, the fundamental challenge remains: the scarcity of high-quality, diverse speech corpora that capture the full spectrum of natural human communication \cite{mihajlik-etal-2024-spoken}.

Hungarian, with its morphologically rich and agglutinative nature, poses a significant challenge even for large multilingual models. To improve the results of these models via fine-tuning, or by training our own, having high-quality spontaneous speech as well as conversational datasets is vital \cite{spont_hun2010}. BEA \citelanguageresource{inproceedings} is a large  Hungarian speech database with the aim, among others, to provide material for research purposes in various fields. The complete 300-hour dataset consists of eight types of speech sessions, including repeated and spontaneous speech, from almost 500 speakers of varying age, gender, and educational background. 

Although the BEA Base dataset \citelanguageresource{mihajlik-etal-2022-bea}, comprising 140 speakers, has been released as a benchmark for ASR of spontaneous Hungarian, a considerable part of the BEA dataset has not yet been processed and subjected to benchmarking. 

In this work, we leverage the remaining BEA recordings to create two new datasets and provide comprehensive baselines for Hungarian conversational ASR. The first dataset, BEA-Large, extends and refines BEA-Base by including additional training data and more extensive segment-level metadata. BEA-Large thus offers an extended corpus of spontaneous Hungarian, with fine-grained annotations to support research in ASR and related areas. The second dataset, BEA-Dialogue, mainly contains dialogues from BEA-Large. Alongside the aforementioned datasets, we provide baseline ASR results for both, using publicly available models to ensure reproducibility. These resources not only advance Hungarian speech technology but also provide a methodological framework for similar efforts in other low-resource languages. By releasing these datasets along with comprehensive baselines, we aim to catalyze research in spontaneous Hungarian speech recognition while contributing to the broader goal of democratizing speech technology across linguistic boundaries.

Our primary contributions are:
\begin{itemize}
	\item An extended dataset (BEA-Large) comprising 255 hours of speech from 433 speakers, substantially expanding the available Hungarian spontaneous speech training data with enriched metadata
	\item A specialized conversational speech dataset (BEA-Dialogue) featuring 85 hours of natural dialogues, addressing the critical shortage of Hungarian dialogue data for conversational ASR and speaker diarization research
	\item Systematic benchmarking using publicly available ASR models to ensure reproducibility and establish performance baselines
\end{itemize}

The structure of the paper is as follows. Section 2 reviews related work. Sections 3 and 4 describe the construction of the two datasets, followed by ASR baseline results in Sections 5 and 6, and diarization results on BEA-Dialogue in Section 7. Finally, Section 8 summarizes the findings and concludes the paper.

\section{Related work}
Concerning the international situation of spontaneous speech datasets, there are several resources for a range of languages and domains. However, a significant portion of these corpora remains unavailable for public use, especially for low-resource languages. Among the publicly accessible English datasets, SSSD \citelanguageresource{sheikh25_interspeech} is designed for dialogue research and features 727 hours of spontaneous English conversations between randomly matched speaker pairs. CASPER \citelanguageresource{xiao2025casperlargescalespontaneous} provides over 100 hours of unscripted English dialogues.

Beyond English, the GRASS corpus \citelanguageresource{schuppler-etal-2014-grass}—the only existing resource for Austrian German for years \cite{linke-etal-2022-conversational}—which contains approximately 19 hours of dyadic speech; and ES-Port \citelanguageresource{garcia-sardina-etal-2018-es}, comprising 40 hours of Spanish dialogues from technical support calls. The RAMC corpus \citelanguageresource{yang2022opensourcemagicdataramcrich} encompasses 180 hours of conversational speech recorded via telephone channels. The Verbmobil dialogue corpus \citelanguageresource{weilhammer-etal-2002-multi} covers multilingual data in German, English, and Japanese. The Kiel Corpus of Spoken German \citelanguageresource{Kohler2017} includes over 8 hours of spontaneous speech from 64 speakers.

For specialized domains, RescueSpeech \citelanguageresource{sagar2023rescuespeechgermancorpusspeech} focuses on the search and rescue context, containing approximately 2 hours of annotated German speech from simulated exercises. The SXUCorpus \citelanguageresource{herms-etal-2016-corpus} offers spontaneous speech components in the Upper Saxon German dialect. The Portuguese CORAA corpus \citelanguageresource{junior2021coraalargecorpusspontaneous} comprises at least 189 hours of spontaneous speech from conversations, monologues, dialogues, and interviews. The CSJ corpus \citelanguageresource{article}, while primarily consisting of monologues, also features spontaneous speech. The Norwegian Parliamentary Speech Corpus \citelanguageresource{solberg2022norwegianparliamentaryspeechcorpus} includes unscripted parliamentary sessions, while the Korean Corpus of Spontaneous Speech \citelanguageresource{article_seul} contains recordings from interviews with 40 speakers.

As discussed by \citet{mihajlik-etal-2024-spoken}, despite the relatively large cumulative volume of Hungarian speech data, inconsistencies in annotation practices and restricted accessibility hinder their simultaneous use—both for general-purpose ASR and for specialized tasks such as conversational ASR. Among the monolingual Hungarian resources, BUSZI ($\sim$600h, 250 spk; \citelanguageresource{buszi}) and SzöSzi (370h, 163 spk; \citelanguageresource{kontra2016}) stand out as extensive sociolinguistic interview collections -- unfortunately, their access is strongly limited. HuComTech ($\sim$50h, 112 spk; \citelanguageresource{papay2011}) provides valuable audiovisual recordings for multimodal communication studies, while BEKK (20h, 56 spk; \citelanguageresource{bodo2017}) contributes spontaneous, student-recorded conversations.

\section{Construction of BEA-Large}
The \emph{BEA-Large} dataset builds upon the original \emph{BEA-Base} corpus by extending its training portion with an additional subset, \emph{BEA-Extension} (later referred to as \emph{train-293}), with each segment corresponding to a single utterance of a given speaker. The \textit{dev} and \textit{eval} partitions remain identical to those in BEA-Base to preserve consistency and comparability across experiments.

\begin{table}[h!]
\centering
\renewcommand{\arraystretch}{1.2}
\setlength{\tabcolsep}{8pt}
\begin{tabular}{l|c|c}
\toprule
& \textbf{train-114} & \textbf{train-293} \\
\midrule
\textbf{\# speakers [f|m]}& 69 | 45  & 184 | 109 \\
\textbf{\# segments} & 69,176  & 196,981  \\
\textbf{\# words} & 555,322 & 1,622,151\\
\textbf{\# chars} & 3,310,493 &  9,550,276 \\
\textbf{duration [h]} & 67.95  & 177.4 \\
\bottomrule
\end{tabular}
\caption{\centering Metadata for the disjoint training sets of BEA-Large.}
\label{tab:large_meta}
\end{table}

The additional training set introduces new metadata attributes that were absent or incomplete in BEA-Base, including speaker \textit{age}, \textit{gender}, \textit{occupation}, and the \textit{module} (e.g., interview, discourse) from which each segment originates. In addition to the unique identifier of the target speaker associated with each recording, the dataset now explicitly labels each speaker's role according to the BEA-Base taxonomy: \emph{SPK} (target speaker), \emph{EXP} (experiment leader), or \emph{DP} (discourse partner).

The \textit{train-293} subset was compiled from recordings of 293 target speakers who do not appear in BEA-Base, ensuring that both training sets can be seamlessly combined without overlap. As summarized in Table~\ref{tab:train-ext_composition}, data were drawn from multiple modules detailed by \citet{mihajlik-etal-2022-bea}, with the \textit{repeat} and \textit{readsent} modules excluded to maintain structural alignment with the \textit{train-114} set. Acoustic preprocessing followed the same protocol as in BEA-Base. 

\begin{table}[h!]
\centering
\renewcommand{\arraystretch}{1.0}
\setlength{\tabcolsep}{10pt}
\resizebox{0.95\columnwidth}{!}{
\begin{tabular}{l|ccc}
\toprule
\textbf{Module} & \textbf{SPK [\%]} & \textbf{EXP [\%]} & \textbf{DP [\%]} \\
\midrule
repeat & -- & -- & -- \\
readsent & -- & -- & -- \\
interview & 16.45 & 3.53 & -- \\
opinion & 12.86 & 5.56 & -- \\
summhist & 3.80 & 0.36 & -- \\
summplant & 3.26 & 0.46 & -- \\
discourse & 21.73 & 18.02 & 8.39 \\
readtext & 5.46 & 0.14 & -- \\
\bottomrule
\end{tabular}}
\caption{\centering Composition of the train-293 set (in percentage of total duration).}
\label{tab:train-ext_composition}
\end{table}

\begin{figure}[h!]
\begin{center}
\includegraphics[height=5cm]{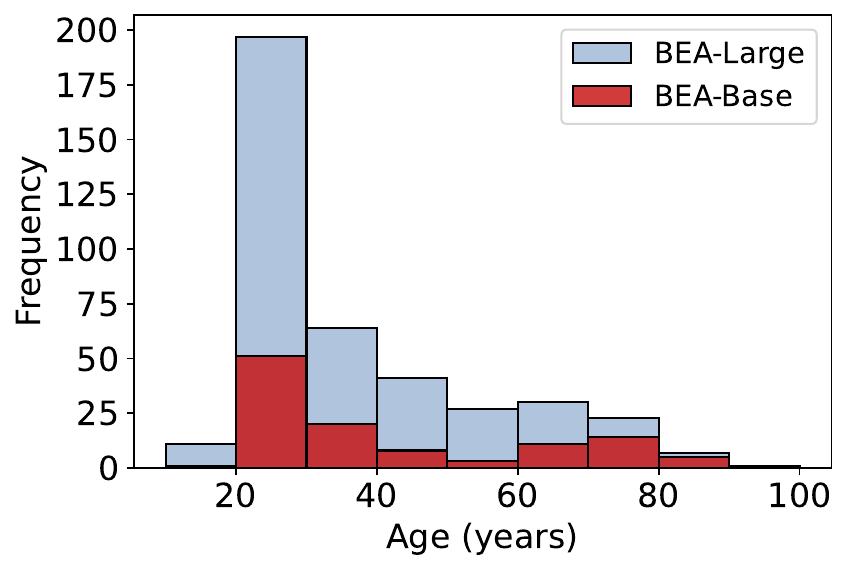}
\end{center}
\caption{\centering Comparison of the age distributions in the train sets of BEA-Large and Base.}
\label{fig:beal-large_agedistr}
\end{figure}

\begin{figure}[h!]
\begin{center}
\includegraphics[height=5cm]{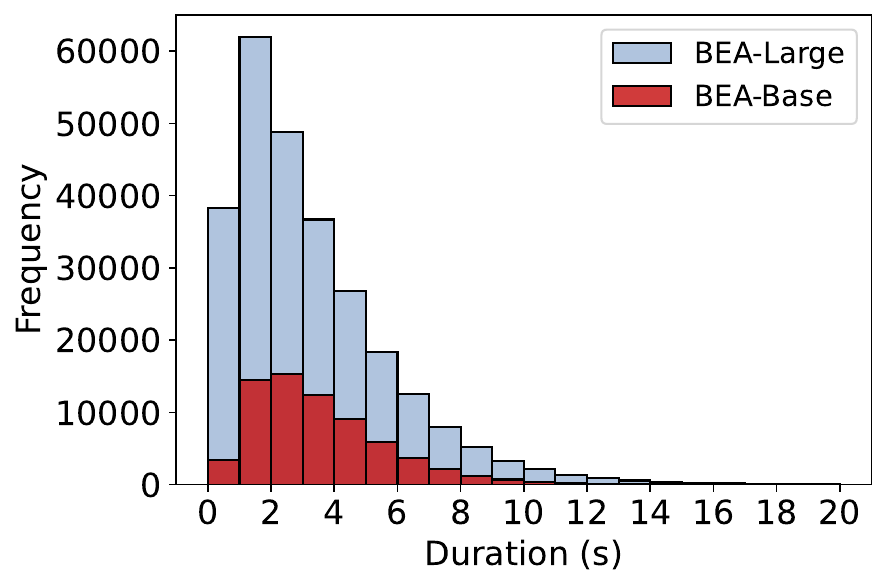}
\end{center}
\caption{\centering Duration of the segments in the train sets of BEA-Large and Base.}
\label{fig:beal-large_durdistr}
\end{figure}

When combined with \textit{train-114}, the resulting BEA-Large training corpus contains approximately three times as much data as BEA-Base. 
The duration of the segments in BEA-Large is shown in Figure \ref{fig:beal-large_durdistr}, with approximately 99.6\% of the segments in the database have a duration of less than 20 seconds.
Figure~\ref{fig:beal-large_agedistr} illustrates the age distribution of speakers in the BEA-Large training set compared to that of \textit{train-114}.

\begin{table*}
\centering
\renewcommand{\arraystretch}{1.2}
\setlength{\tabcolsep}{8pt}
\resizebox{2\columnwidth}{!}{
\begin{tabular}{l|c|cc|cc|cc|cc}
\toprule
\multirow{2}{*}{\textbf{Model}} & \multirow{2}{*}{\textbf{Dataset}} & \multicolumn{2}{c|}{\textbf{dev-repet}} & \multicolumn{2}{c|}{\textbf{eval-repet}} & \multicolumn{2}{c|}{\textbf{dev-spont}} & \multicolumn{2}{c}{\textbf{eval-spont}} \\ 
\cmidrule(lr){3-4} \cmidrule(lr){5-6} \cmidrule(lr){7-8} \cmidrule(lr){9-10}
 &  & \textbf{WER} & \textbf{CER} & \textbf{WER} & \textbf{CER} & \textbf{WER} & \textbf{CER} & \textbf{WER} & \textbf{CER} \\
\midrule
\Verb|whisper-large-v3| & \textit{zero-shot} &\textit{\textbf{13.31}} & \textit{\textbf{2.74}} & \textit{\textbf{13.73}} & \textit{\textbf{2.83}} & \textit{\textbf{20.79}} & \textit{\textbf{9.22}} & \textit{\textbf{21.28}} & \textit{\textbf{9.31}} \\ 
\Verb|whisper-large-v2| & \textit{zero-shot} &  \textit{18.56} & \textit{4.47} & \textit{21.26} & \textit{4.46} & \textit{34.71} & \textit{19.67} &  \textit{32.56} & \textit{18.61} \\ 
\Verb|whisper-medium| & \textit{zero-shot} &\textit{ 22.7} & \textit{5.27}& \textit{23.84} & \textit{6.10} & \textit{37.23} & \textit{20.13} & \textit{39.19} & \textit{21.71}\\ 
\midrule
\multirow{3}{*}{\Verb|f\_conformer\_ctc|} & train-114 & 7.01 & 1.54 &  7.05 & 1.82 & 18.26 & 6.04 & 19.22 & 6.37\\
                      & train-293 & 14.60 & 2.83 & 14.98 & 3.30 &15.60 & 5.18 &16.33 & 5.34\\
                     &train-114 + train-293& \textbf{5.21} & \textbf{1.16} & \textbf{4.80} & \textbf{1.20} & \textbf{13.52} & \textbf{4.41} & \textbf{14.18} & \textbf{4.56}\\


\bottomrule
\end{tabular}}
\caption{\centering WER and CER [\%] for zero-shot and fine-tuned models on BEA-Large under different configurations.}
\label{tab:bl_fc_results}
\end{table*}

\section{Construction of BEA-Dialogue}

A dedicated dialogue corpus, \emph{BEA-Dialogue}, was derived from the recordings of 242 speakers that had not previously been included in BEA-Base.

In the original dataset, utterances were organized by the target speaker, with each module stored in a \texttt{TextGrid} file containing time-aligned transcriptions for all speakers appearing in that module. To create BEA-Dialogue, utterances were first extracted along with their timestamps and speaker labels (\emph{SPK}, \emph{EXP}, \emph{DP}). 

\begin{figure}[h!]
\begin{center}
\includegraphics[width=1.0\linewidth]{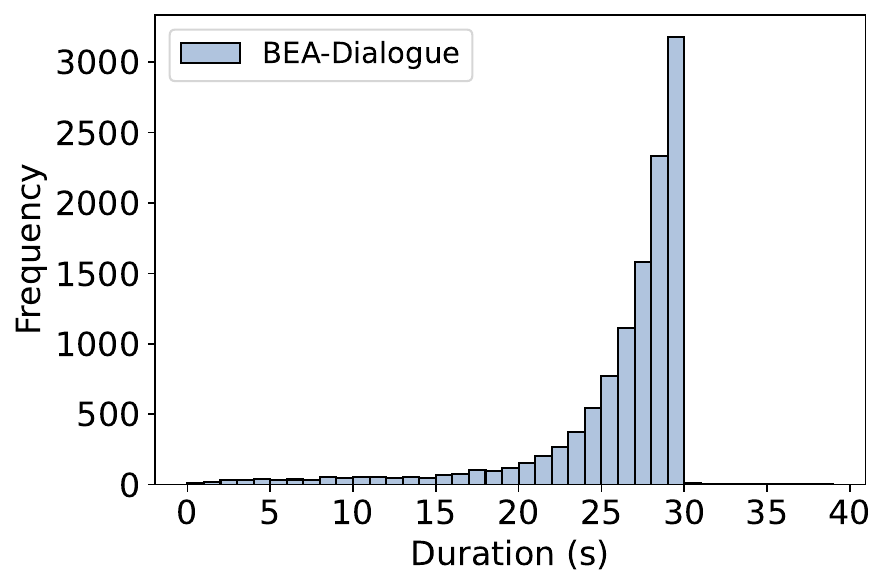}
\end{center}
\caption{\centering Duration of the segments in BEA-Dialogue.}
\label{fig:bea_dial_duration}
\end{figure}

Candidate cut points for shorter dialogue segments were then identified by detecting silent intervals—regions that did not overlap with any utterance, except possibly at their boundaries. Using these silence-based boundaries, utterances from multiple speakers were grouped into coherent dialogue units according to the silent regions separating them. These smaller units were subsequently merged into larger dialogue segments, each with a target duration of 30 seconds. Figure \ref{fig:bea_dial_duration} shows the histogram of the durations after the procedure.

Beyond the target speakers, the dataset includes the voices of five additional female speakers (\textit{fem1--fem5}) and two male speakers (\textit{male1--male2}), who served as experiment leaders and discourse partners. To construct \emph{BEA-Dialogue}, which provides fully disjoint training, development, and evaluation sets across all speaker roles, the data were partitioned based on three experiment leaders: \textit{fem1}, \textit{fem3}, and \textit{fem4}. 

Since these experiment leaders also acted as discourse partners for different target speakers, we excluded dialogue segments from the discourse module in which the target speaker’s discourse partner was one of the three experiment leaders. The combinations of experiment leaders and discourse partners across all available data are shown in Table \ref{tab:exp_dp_pairs}, along with the corresponding number of dialogue segments belonging to each pair. 

\begin{table}[h!]
\centering
\renewcommand{\arraystretch}{1.0}
\resizebox{0.7\columnwidth}{!}{
\setlength{\tabcolsep}{8pt}
\begin{tabular}{c|c|ccc}
\toprule
 &  & \multicolumn{3}{c}{\textbf{EXP}} \\ 
\cmidrule(lr){3-5}
 &  & \textbf{fem1} & \textbf{fem3} & \textbf{fem4}  \\ 
\midrule
\multirow{7}{*}{\textbf{DP}}
 & \textbf{fem1} & -- & 286  & 852  \\ 
 & \textbf{fem2} & 1751  & -- & 122  \\ 
 & \textbf{fem3} & 4415  & -- & 546  \\ 
 & \textbf{fem4} & 4418  & -- & --  \\ 
 & \textbf{fem5} & 1662  & 367  & 816  \\ 
 & \textbf{male1} & 983  & 88  & 588  \\ 
 & \textbf{male2} & 46 & -- & --  \\ 
\bottomrule
\end{tabular}}
\caption{\centering Number of discourse segments by experiment leader (EXP) and discourse partner (DP) across all dialogue segments.}
\label{tab:exp_dp_pairs}
\end{table}

The training set includes dialogue segments where the experiment leader is \textit{fem1} and the discourse partner is either \textit{fem2} or \textit{male2}, as well as all segments from other modules featuring \textit{fem1}. The development and evaluation sets contain dialogues where the experiment leaders are \textit{fem3} and \textit{fem4}, with discourse partners \textit{fem5} and \textit{male1}, respectively. Table~\ref{tab:dial_meta} summarizes the metadata of BEA-Dialogue, which represents the largest set that can be derived from the data without violating speaker independence across subsets with respect to the three speaker roles (SPK, EXP, DP). 


\begin{table}[h!]
\centering
\renewcommand{\arraystretch}{1.2}
\setlength{\tabcolsep}{8pt}
\resizebox{1\columnwidth}{!}{
\begin{tabular}{l|ccc}
\toprule
& \multicolumn{3}{c}{\textbf{BEA-Dialogue}} \\
\cmidrule(lr){2-4} 
& \textbf{Train} & \textbf{Dev} & \textbf{Eval}  \\
\midrule
\textbf{\# Speakers [f|m]} & 121 | 67 & 3 | 6 & 29 | 16 \\
\textbf{\# Segments} & 9{,}179 & 577 & 1{,}906\\
\textbf{\# Words} & 532{,}732 & 34{,}056 & 105{,}472  \\
\textbf{\# Characters} & 3{,}217{,}617 & 206{,}740 & 641{,}628 \\
\textbf{Avg. \# Speakers / Seg} & 1.77 & 1.92 & 1.61 \\
\textbf{Avg. \# Utterances / Seg} & 10.99 & 8.68 & 9.74  \\
\textbf{Avg. Seg Duration [s]} & 26.23 & 26.31 & 26.09  \\
\textbf{Overlap Duration [h]} & 3.28 & 0.29 & 0.40  \\
\textbf{Total Duration [h]} & 66.87 & 4.22 & 13.81 \\
\bottomrule
\end{tabular}}
\caption{\centering Metadata of the BEA-Dialogue dataset.}
\label{tab:dial_meta}
\end{table}

\section{Baseline ASR results for BEA-Large}

To establish reproducible baselines, we fine-tuned a potent yet relatively small (120M parameters) model publicly available in the NeMo toolkit \cite{kuchaiev2019nemotoolkitbuildingai} -- \verb|STT En Fast Conformer-CTC Large| \footnote{\url{https://huggingface.co/nvidia/stt_en_fastconformer_ctc_large}} (\verb|f_conformer_ctc|) 
-- on three different training sets: \textit{train-114 (BEA-Base)}, \textit{train-293 (BEA-Extension)} and \textit{train-114 + train-293 (BEA-Large)}.  


Performance was evaluated using two standard metrics: Word Error Rate (WER) and Character Error Rate (CER). The results are summarized in Table~\ref{tab:bl_fc_results}. Zero-shot benchmarks obtained using the Speechbrain toolkit \cite{speechbrain, speechbrain_v1} with three Whisper models\footnote{\url{https://github.com/openai/whisper}} --- \texttt{whisper-large-v3}, \texttt{whisper-large-v2}, and \texttt{whisper-medium} \cite{radford2022robustspeechrecognitionlargescale} --- are also presented in Table~\ref{tab:bl_fc_results}. 

The Fast Conformer model fine-tuned on less than 250 hours of Hungarian speech consistently outperformed all Whisper models, across both the repetitive and spontaneous subsets of BEA-Large. 

In both training partitions of BEA-Large, the voices of the seven individuals acting as experiment leaders and discourse partners are notably overrepresented. To obtain two speaker-independent subsets, we extracted utterances from the two training sets belonging exclusively to the target speakers.
We constructed three setups from the two speaker-independent supplementary training sets: using each dataset individually, one at a time, and using both datasets merged. Notably, the WER and CER increased when the training sets were limited to the utterances of the target speakers. 

Using the new train-293 as the only training set resulted in higher WER and CER on dev-repet and eval-repet compared to using train-114. This increase may be caused by data leakage between the original BEA-Base splits, which will be the subject of further investigation.

The best results were achieved by using the two full components -- train-114 and train-293 -- of the database yielding a WER of 14.18\% and a CER of 4.56\% on the most important spontaneous evaluation subset, indicating that using the voice of the experiment leaders and discourse partners does not have a negative impact on the evaluation results obtained on a speaker-independent evaluation set.

\section{Baseline ASR results for BEA-Dialogue}
For the \textit{BEA-Dialogue} dataset, we trained a similar set of models to those used for \textit{BEA-Large}, complemented with fine-tuned Fast Conformer models. During training, we employed Serialized Output Training (SOT)~\cite{SOT}, marking speaker transitions explicitly with a \texttt{<sc>} (speaker change) token. These tokens were inserted to reflect the structure of an ideal dialogue transcript. For example (English translations in parentheses):

\begin{quote}
\textit{Hogy vagy?} \texttt{<sc>} \textit{Köszönöm, jól. És te?} \texttt{<sc>} \textit{Én is, köszönöm.} \texttt{<sc>} \textit{Örülök neki.} 
(\textit{How are you?} \texttt{<sc>} \textit{I'm good, thanks. And you?} \texttt{<sc>} \textit{Me too, thanks.} \texttt{<sc>} \textit{I'm glad.})
\end{quote}

The utterance of each speaker is maintained as an uninterrupted sequence, even when there is overlap in speech. This allows the model to learn to recognize speaker boundaries while preserving the linguistic integrity of each turn.

Evaluation followed the same criteria as in BEA-Large: \textit{WER} and \textit{CER}, complemented by the \textit{concatenated minimum-permutation WER} (cpWER) and \textit{cpCER}, which minimize errors across all possible permutations of speakers separated by \texttt{<sc>} tokens. Since some segments contained more than ten speaker changes, exhaustive permutation computation was infeasible. To address this, we adopted a hybrid strategy: computing all permutations for up to seven speaker changes and applying beam search to achieve near-optimal results for higher counts. For fine-tuned models, we additionally report the \textit{speaker change accuracy} (\textit{scAcc})—the proportion of utterances in which the predicted number of speaker transitions matches the reference (i.e., occurrences of \texttt{<sc>} tokens).

Table~\ref{tab:bea_dialogue_results} summarizes the obtained results.

\begin{table*}[t]
\centering
\resizebox{\textwidth}{!}{
\begin{tabular}{l|ccccc|cccccc}
\toprule
\multirow{2}{*}{\textbf{Model}} & \multicolumn{5}{c}{\textbf{dev}} & & \multicolumn{5}{c}{\textbf{eval}} \\
\cmidrule(lr){2-6} \cmidrule(lr){8-12}
 & \textbf{WER} & \textbf{cpWER} & \textbf{CER} & \textbf{cpCER} & \textbf{scAcc} & & 
   \textbf{WER} & \textbf{cpWER} & \textbf{CER} & \textbf{cpCER} & \textbf{scAcc} \\
\midrule
\texttt{whisper-large-v3}   & 21.19 & 21.04 & 12.74 & 12.56 & -- & & 22.21 & 22.13 & 12.27 & 12.18 & -- \\
\texttt{whisper-large-v2}   & \textbf{19.65} & \textbf{19.42} & 9.84  & 9.58  & -- & & 24.50 & 24.42 & 13.13 & 13.05 & -- \\
\texttt{whisper-medium}     & 25.45 & 25.27 & 12.61 & 12.42 & -- & & 29.21 & 29.12 & 14.71 & 14.63 & -- \\
\midrule
\texttt{f\_conformer\_ctc}     & 19.69 & 19.53 & \textbf{7.95} & \textbf{7.78} & 69.32 & & \textbf{20.56} & \textbf{20.44} & \textbf{9.11} & \textbf{9.00} & 82.16 \\
\bottomrule
\end{tabular}}
\vspace{5pt}
\caption{ASR accuracy (\%) comparison on BEA-Dialogue.}
\label{tab:bea_dialogue_results}
\end{table*}

As shown in Table~\ref{tab:bea_dialogue_results}, the fine-tuned \texttt{f\_conformer\_ctc} model achieves lower error rates on the \textit{eval} set, demonstrating the benefits of domain-specific adaptation. Interestingly, however, on the \textit{dev} set, the \texttt{Whisper-large-v2} model outperforms both its \texttt{v3} variant and the fine-tuned model. Nevertheless, the fact that an English model fine-tuned on fewer than 70 hours of Hungarian data can surpass state-of-the-art systems underscores the importance of corpora such as BEA-Dialogue for advancing Hungarian conversational ASR. These findings highlight the current limitations of state-of-the-art models when applied to spontaneous, real-world speech.

The ratio of character error rate (CER) to word error rate (WER) is notably higher than typically observed in non-conversational datasets, suggesting that fine-grained, character-level deviations occur more frequently in spontaneous dialogue.

\begin{figure}[h!]
\centering
\includegraphics[width=1.0\linewidth]{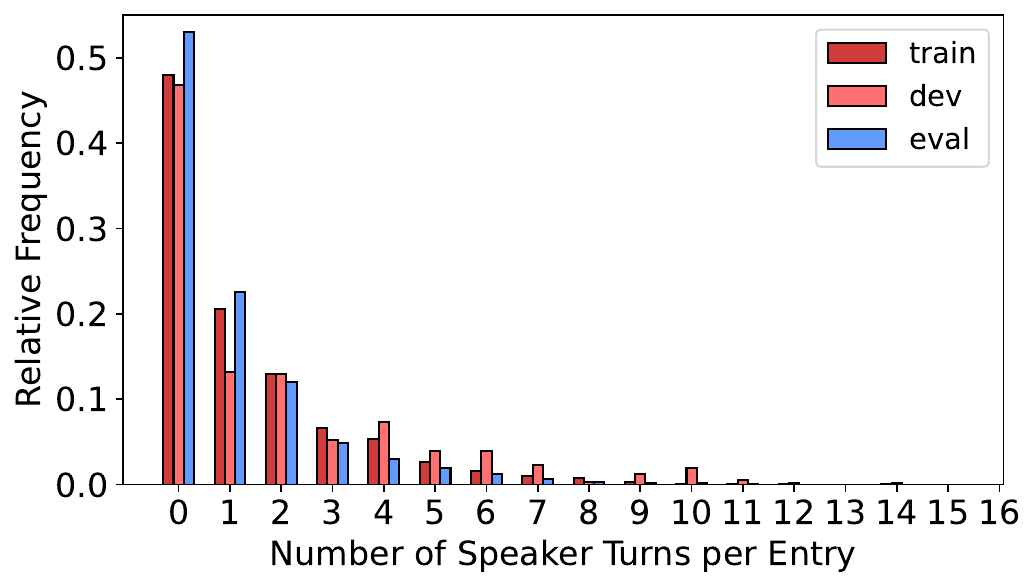}
\caption{\centering Distribution of speaker changes per segment in BEA-Dialogue.}
\label{fig:segment_distribution}
\end{figure}

Figure~\ref{fig:segment_distribution} shows the distribution of the number of speaker turns per segment for each split. As shown, most segments contain no or only one speaker change, making the task largely comparable to single-speaker spontaneous speech recognition. However, the ones with a higher number of turns substantially increase task complexity.

\section{Baseline diarization results for BEA-Dialogue}
We establish baseline diarization performance for the BEA-Dialogue dataset using its \textit{dev} and \textit{eval} splits. The goal of these baselines is to provide reference performance levels for future work on speaker segmentation.

Two state-of-the-art diarization systems were evaluated: \textit{pyannote.audio}~\cite{pyannote} and \textit{Sortformer}~\cite{sortformer}. Both models were used in their pre-trained configurations without additional fine-tuning. The \textit{pyannote.audio} model represents a hybrid neural approach that combines segmentation and clustering stages, while \textit{Sortformer} adopts a transformer-based end-to-end architecture with a sorting-based loss function to better handle overlapping speech. The exact pre-trained checkpoints used are: \texttt{speaker-diarization-3.1}\footnote{\url{https://huggingface.co/pyannote/speaker-diarization-3.1}} and \texttt{diar\_sortformer\_4spk-v1}\footnote{\url{https://huggingface.co/nvidia/diar_sortformer_4spk-v1}}.

Table~\ref{tab:der_comparison} reports the average Diarization Error Rate (DER) obtained by each model on the two dataset subsets.

\begin{table}[h!]
\centering
\begin{tabular}{lcc}
\toprule
\multirow{2}{*}{\textbf{Model}} & \multicolumn{2}{c}{\textbf{DER (\%)}} \\ 
\cmidrule(lr){2-3}
 & \textbf{dev} & \textbf{eval} \\ 
\midrule
Pyannote & 14.09 & 17.40 \\
Sortformer & \textbf{12.46} & \textbf{15.11} \\
\bottomrule
\end{tabular}
\caption{\centering Average DER comparison on BEA-Dialogue.}
\label{tab:der_comparison}
\end{table}

\begin{figure}[h!]
\centering
\includegraphics[width=0.9\linewidth]{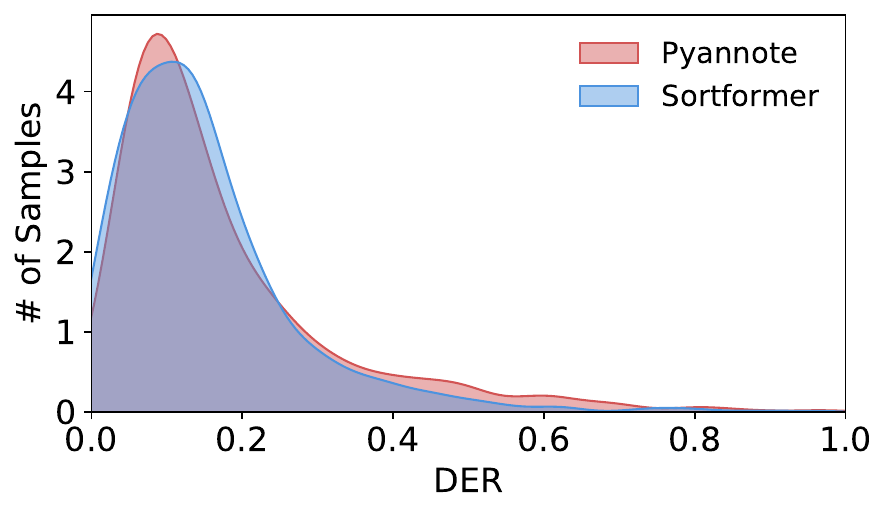}
\caption{\centering DER distribution on BEA-Dialogue (eval).}
\label{fig:der_distribution}
\end{figure}

Figure~\ref{fig:der_distribution} illustrates the distribution of DER values across individual dialogue segments in the \textit{eval} subset. Both systems exhibit similar overall trends.

\section{Conclusion}
This work addresses the shortage of high-quality spontaneous and conversational speech data for Hungarian by introducing two substantial resources: BEA-Large and BEA-Dialogue.
BEA-Large extends the original BEA-Base corpus with approximately 178 hours of additional speech from 293 speakers, enriched with segment-level metadata such as age, gender, occupation, and speaker role. This expansion nearly triples the amount of available training data while maintaining compatibility with the original benchmark splits.

BEA-Dialogue, comprising 85 hours of natural conversations, represents one of the largest Hungarian datasets explicitly designed for conversational ASR and speaker diarization research. The dataset includes predefined splits that ensure complete speaker independence across all conversational roles.

Our baseline experiments yield several key findings. Fine-tuned Fast Conformer models achieved WERs as low as 14.18\% on spontaneous speech and 4.8\% on repeated speech when trained on the combined datasets, highlighting the benefits of including data from both experiment leaders and discourse partners. In conversational ASR, the relatively high CER-to-WER ratio compared to monologue speech suggests that spontaneous dialogue introduces additional challenges—arising from phenomena such as overlapping speech, rapid turn-taking, and interaction-driven disfluency patterns that are not present in single-speaker settings. The speaker diarization baselines, with DERs ranging from 12.46\% to 17.40\%, establish solid reference points for future advancements.

By releasing these datasets together with reproducible baselines built on publicly available models, we aim to accelerate research in Hungarian speech technology and provide a methodological blueprint for similar initiatives in other low-resource languages. Future work will focus on advanced methods for overlap handling, speaker change detection, and context-aware modeling, with the goal of improving both ASR and diarization performance in conversational scenarios.

\section*{Acknowledgment}
Project No. 2025-2.1.2-EKÖP-KDP-2025-00005 has been implemented with the support provided by the Ministry of Culture and Innovation of Hungary from the National Research, Development and Innovation Fund, financed under the EKÖP\_KDP-25-1-BME-21 funding scheme.

The work was also partially supported by the Hungarian NRDI Fund through the projects NKFIH K143075 and K135038, NKFIH-828- 2/2021(MILAB).





\section{Bibliographical References}\label{sec:reference}

\bibliographystyle{lrec2026-natbib}
\bibliography{main}

\section{Language Resource References}
\bibliographystylelanguageresource{lrec2026-natbib}
\bibliographylanguageresource{resource_lang}

\end{document}